\documentclass[10pt, a4paper]{article}

\usepackage[final]{scrable} 

\title{Self-Improving Customer Review Response Generation Based on LLMs}

\name{Guy Azov*, Tatiana Pelc*, Adi Fledel Alon\textsuperscript{\textdagger}, Gila Kamhi\textsuperscript{\textdagger}\thanks{*Equal contribution}\thanks{\textdagger Joint supervision}}

\address{Intel Corporation}

\abstract{
Previous studies have demonstrated that proactive interaction with user reviews has a positive impact on the perception of app users and encourages them to submit revised ratings. Nevertheless, developers encounter challenges in managing a high volume of reviews, particularly in the case of popular apps with a substantial influx of daily reviews. Consequently, there is a demand for automated solutions aimed at streamlining the process of responding to user reviews. To address this, we have developed a new system for generating automatic responses by leveraging user-contributed documents with the help of retrieval-augmented generation (RAG) and advanced Large Language Models (LLMs). Our solution, named SC\textbf{R}ABLE,  represents an adaptive customer review response automation that enhances itself with self-optimizing prompts and a judging mechanism based on LLMs. Additionally, we introduce an automatic scoring mechanism that mimics the role of a human evaluator to assess the quality of responses generated in customer review domains. Extensive experiments and analyses conducted on real-world datasets reveal that our method is effective in producing high-quality responses, yielding improvement of more than $8.5\%$ compared to the baseline. Further validation through manual examination of the generated responses underscores the efficacy our proposed system.
 \\ \newline  }

\begin{document}

\maketitleabstract

\section{Introduction}
\label{sec:intro}
Large language models (LLMs) have demonstrated remarkable performance in a wide range of tasks related to comprehending and generating natural language, text, and code \cite{devlin2019bert,2019arXiv191010683R,brown2020language}, \cite{zhang2022opt, chowdhery2022palm, chung2022scaling}. 
The most notable advancement is that these tasks are executed using few-shot or in-context learning\cite {xie2022explanation,dong2023survey, roberts2020knowledge}, reducing the need for construction of traditional labeled datasets for supervised learning. Through their ability to efficiently store and apply knowledge, LLMs have shown outstanding capabilities in tasks involving information-seeking questions, where the question cannot be answered easily by the person asking it \cite{tunstall2022efficient}. Large Language Models (LLMs) are advanced AI systems designed to understand, generate, and manipulate human language. They are trained on vast amounts of text data, allowing them to perform a wide range of language-related tasks. In the contemporary digital age, customer reviews have become a cornerstone of consumer decision-making. Prospective buyers frequently rely on online reviews as a principal source for obtaining insights into various products and services. Empirical research indicates a robust and positive correlation between the numerical rating of a mobile application and the number of downloads it garners. Furthermore, it has been observed that users exhibit a tendency to modify their ratings following the reception of responses from developers. Consequently, the act of responding to user reviews is considered imperative in the realm of app development. However, crafting an appropriate response to an online review is a complex task that demands expertise to ensure it matches the customer's feedback in both content and tone. A response must cater to different audiences: the reviewer seeking resolution or acknowledgment, potential customers who use reviews to inform their buying choices, and search engines that use this content for search ranking purposes. The sheer volume and diversity of customer reviews across platforms like e-commerce sites, social media, and review websites present both a treasure trove of information and a daunting challenge for consumers seeking answers and for businesses. The latter often struggle with the resources and time to manage this feedback effectively, and they may not have staff skilled in crafting responses. In this paper, we introduce a scalable automatic end-to-end customer review response generation methodology based on LLMs. We  aim to get high-quality responses leveraging  an optimization strategy that  relies on \textit{LLM-as-a-Judge}, capable of iteratively  scoring and proposing response improvements. Subsequently, these proposals are fed into a prompt generator that generates an improved prompt for each iteration in the response generation process.
Our methodology, which deploys custom-tailored prompts for every customer support category, has demonstrated superior performance over the general prompt as per the research conducted by  \cite{yuan2024selfrewarding}.
\paragraph{Overall, we make the following contributions:}
\begin{itemize}
\item  We propose (SC\textbf{R}ABLE - Self-Improving Customer Review Response Automation Based on LLMs), an LLM-based approach to automatically generate high quality responses to given user reviews. We demonstrate the power of customized prompt engineering to lead the LLM-based solutions to  responses that raise customer satisfaction, engagement and delight. Furthermore, we employ automatic prompt engineering, using an LLM to improve a prompt, which is then evaluated against an objective function evaluator. We achieve an optimal review response prompt for inference via a two step method,  1) Review - Response generating LLM (calibrated by human evaluation) 2) Automatic prompt optimization using \textit{LLM-as-a-Judge}.
\item We conduct both manual and automatic evaluation on the performance of the proposed models and baselines. The experimental results indicate that our optimized prompt increased the human score of our test set response generations by more than $\mathbf{8.5\%}$ comapred to the generations obtained by using our initial base prompt.
\item The results demonstrate that our proposed \textit{LLM-as-a-Judge} approach achieves ~3-5 times stronger correlation with human evaluation compared to \cite{yuan2024selfrewarding}.
\end{itemize}
The rest of this paper is organized as follows. Section \ref{sec:related} surveys the related work. Section \ref{sec:pipeline} introduces an overview of the proposed approach and the detailed design of the approach. Section \ref{sec:results}  elaborates on the experimental results, including the results from the automatic  LLM based evaluation and manual human evaluation. Sections \ref{sec:discussion} and \ref{sec:conclusions} discuss conclude our work, summarizing the proposed future work.
\section{Related Work}\label{sec:related}
\subsection{Customer Reviews Analysis}
As noted in Pagano et al., user feedback and user involvement are crucial for modern software organizations \cite{Pagano2013}. Data mining of user reviews has attracted significant research attention owing to the pivotal role reviews play in shaping consumer perceptions and decision-making regarding applications. Researchers have applied various techniques to analyze these reviews, ranging from fundamental structural features, such as review length and TF-IDF (Term Frequency-Inverse Document Frequency), which are frequently used to automatically classify user review emotions at a high level. Furthermore, more in-depth analyses have been pursued through the extraction of content features, including sentiment, topic, and keywords, often achieved through the application of machine learning methods\cite {6912257,7765038,8804432,7985654,bharti2017automatic}. Other papers provide a unified summary of multiple customer reviews using machine learning models \cite {brazinskas-etal-2020-shot,brazinskas2022efficient,bhaskar2023prompted}.
\subsection{Customer Reviews Response Generation}
In addition to the process and analysis of the reviews, it is crucial to properly respond to the user. In addition to being informative, such response should be polite, address user's concerns, be empathic, leave a positive impression about the product, etc. It is important for developers to carefully respond to  each and every customer review. Hassan et al. indicate that the chances that a user will revisit their review score are six times higher if the review gets a timely and to-the-point response from the product team \cite {hassan2018studying}. However, some applications have so many users and reviews such that human responses are not always possible for all of the reviews. In recent years, efforts were made to automatically generate responses to customer reviews using machine learning techniques. Gao et al. suggest an RNN-based model named \textit{RRGen} to encode the review with high level features such as occurrences of specific keywords, rating, review sentiment, review length and app category towards an automatic response generation \cite{gao2020automating}. Zhang et al.\cite{zhang2023transformer} propose a transformer \cite{vaswani2017attention} based model named \textit{TRRGen} for automatic app review response generation. \textit{TRRGen} fuses the features of app category and ratings and demonstrates that the fusion of app category feature and rating feature into token semantics is helpful for generating high-quality responses (competitive with human app expert responses). Gao et al. aim to address two limitations of the method they previously suggested, namely its lack of flexibility and generalization, which often leads to the generation of non-informative responses \cite{gaoautomating}. Their proposed solution, named CoRe, leverages app details and responses from similar reviews. In addition, Farooq et al. train a seq2seq model with a retrieval component that merges user reviews with pertinent app descriptions and known user reviews, using specific app features to generate app-aware responses \cite{farooq2020app}. Cao et al. evaluate the performance of selected pre-trained language models against a transformer model trained from scratch in the context of automatic customer review response generation. They find that although pre-trained language models may score lower than baseline models in their experiments, they still prove effective in generating responses and show considerable robustness relative to the amount of training data used \cite{cao2022pretrained}.
Finally, Chen et al. propose a multi aspect attentive network to automatically attend different aspects of the review, ensuring most of the issues are being answered \cite{chen2022generating} 
\subsection{Response Evaluation} 
Assessing the quality of generated responses in the context of generative AI models involves multiple parameters such as relevance, coherence, and human-likeness. In the study by Katsiuba et al. \cite{katsiuba2023artificially}, an online experiment involving 502 participants was leveraged to determine the effectiveness of large language models (LLMs) in generating responses to customer feedback. The experiment's findings indicate that LLMs' responses were not only effective in achieving communicative goals but also held up well when compared to responses written by humans. One key methodology employed to evaluate the responses was the Turing test approach \cite{turing2009computing}, which involves human evaluators to determine the human-like quality of an utterance generated by an AI. \newline Traditional automatic evaluation metrics such as BLEU \cite{papineni2002bleu}, ROUGE \cite{lin2004rouge}, and METEOR \cite{banerjee2005meteor} often do not correlate well with human judgment due to their focus on lexical matching. Consequently, there is a pressing need for more advanced automatic evaluation techniques that better mirror human assessments. One approach is to employ semantic evaluation methods that measure the similarity between the ground truth and model-generated responses \cite{zhang2019bertscore, zhao2019moverscore, risch2021semantic}. Another emerging strategy is to utilize Large Language Models (LLMs) as evaluators to assess the quality of text and the overall performance of language language models, a practice known as \textit{LLM-as-a-Judge} \cite{fu2023gptscore, gao2023human, chiang2023can, liu2023g, shen2023large, wang2023large, wang2023far, peng2023instruction, gudibande2023false, zhou2023lima, dettmers2023qlora, dubois2023alpacafarm, bubeck2023sparks, chan2023chateval, yuan2021bartscore, alpaca_eval, fernandes2023devil, bai2023benchmarking, saha2023branch, kim2023prometheus, zheng2023judging, kim2023evallm}. While the focus has been on the automatic evaluation of responses, the integration of retrieval-augmented generation (RAG) frameworks \cite{lewis2020retrieval, guu2020retrieval, izacard2022few} has become increasingly prevalent to boost LLM performance. This integration necessitates the development of an automated evaluation system tailored for the comprehensive RAG process \cite{es2023ragas, saad2023ares}.
\subsection{LLM Self Improvement}
While Large Language Models (LLMs) are adept at generating content, they may not always cater to specific use case requirements. To tackle this issue, enhancing LLMs through self-improvement techniques has become a focal point of research. Madaan et al. \cite{madaan2023self} introduce \textit{SELF-REFINE}, a technique for the autonomous enhancement of an LLM through cycles of feedback and refinement. Zhou et al. \cite{zhou2022large} present the \textit{Automatic Prompt Engineer}, a method for choosing prompts that optimize a particular score function. Furthermore, Yang et al. \cite{yang2023large} explore the application of LLMs as optimizers with their approach, \textit{Optimization by PROmpting}. Pryzant et al. \cite{pryzant2023automatic} suggest \textit{Prompt Optimization with Textual Gradients}, a non-parametric strategy influenced by gradient descent to fine-tune prompts according to a scoring function. Another study by Wang et al. \cite{wang2023promptagent} views prompt optimization as a form of strategic planning, proposing \textit{PromptAgent} to autonomously generate expert-level prompts. Wang et al. \cite{wang2022self} propose \textit{Self-Instruct}, a method for bootstrapping LLMs using instruction-response pairs that they generate themselves. Lastly, Yuan et al. \cite{yuan2024selfrewarding} investigate Self-Rewarding Language Models, which are capable of self-improvement by evaluating and training on their own outputs. These models not only employ \textit{LLM-as-a-Judge} for self-assessment; but also use training data to create instructions that enhance the quality of the target output. Iterative methods use a single LLM to act as the generator, refiner, and feedback provider or to generate and judge its own responses to improve both its response quality and reward prediction ability.  The \textit{SELF-REFINE} framework allows large language models to iteratively improve their output by generating initial output, evaluating it, and then refining it based on self-generated feedback, all without the need for additional or external data or training. This method harnesses the model's own feedback to enact self-improvement, similar to human revision processes \cite{madaan2023self}. The self-improving process involves Self-Rewarding Language Models (SLMs) starting with a base pre-trained language model and a small amount of human-annotated seed data, which then engage in self-instruction creation to generate and judge new training data \cite{yuan2024selfrewarding}. Each iterative cycle aims to surpass the previous models by using refined training sets from the model's own generations and evaluations, leading to both improved instruction-following abilities and a dynamic, improving reward modeling capacity. Integrating human expertise with AI, in customer feedback management improves the generation of human-like responses. Human-AI collaborative configurations, such as a combination of deep learning models with human edits, showcased better performance in Turing tests, suggesting they were more human-like than responses from AI alone \cite{katsiuba2023artificially}. The significant amplification in communicative effectiveness, offering responses that align more closely with customer expectations in terms of quality, fairness, and personalization. Our approach integrates artificial intelligence to enhance customer review analysis by focusing on key elements like accuracy, relevance, and empathy, essential for the customer support domain. By incorporating the \textit{LLM-as-a-Judge} system, we've introduced an intermediate prompt creation step, which allows for a more controlled and nuanced adjustment process. This strategy involves selectively choosing categories for review and methodically suggesting on which reviews to base new prompts, ensuring a more tailored and impactful response to users. Moreover, our system is designed with stability in mind; the feedback and \textit{LLM-as-a-Judge} mechanisms are fixed, eliminating the need for generating new training data. The collaboration between the judge LLM and the nuanced prompts across different categories delivers more rounded, human-like responses. Additionally, we have implemented an automated scoring method for the model which correlates well with human judgment, ensuring that our automatic assessment and scoring align closely with human perspectives.
\section {Methods}\label{sec:pipeline}
\subsection{LLM as a Judge of Customer Review Response}
Due to the limited availability, the challenge of obtaining, and the expenses related to human evaluations, our aim was to create an automated system, \textit{LLM-as-a-Judge}, that is designed to evaluate customer feedback responses just like a human judge would. Undoubtedly, such a tool provides us with the capability to evaluate online reviews and enhance our services in real-time. It not only facilitates immediate feedback but also paves the way for ongoing enhancements in the way we provide our services. Our approach assumes that for a given collection of customer reviews, denoted as  $\{R_{i}\}_{i=1}^{N}$, there exist corresponding responses crafted by human experts, denoted as $\{\textit{ExpertResponse}[R_{i}]\}_{i=1}^{N}$. These expert responses act as ideal examples, illustrating the optimal reply for each particular review. In developing an \textit{LLM-as-a-Judge} intended to serve as a proxy to for actual human evaluation, we initially requested each author of the paper to evaluate the responses based on four criteria - \textit{Relevancy} - how relevant the response is regarding to the review, \textit{Application Specificity} - how specific the response is regarding the application, at hand, \textit{Accuracy} - how accurate the response is and \textit{Grammatical Correctness} of the response. These criteria were selected because they are widely recognized in the customer review domain (\cite{zhang2023transformer}, \cite{gao2020automating},\cite{gaoautomating}, \cite{farooq2020app}) . Evaluators assign ratings to each aspect individually on a scale from 1 to 5. Drawing on the works of \cite{liu2023g} and \cite{yuan2024selfrewarding}, we devised specific evaluation prompts for each category that reflect the guidelines given to the human responders (detailed prompts are included in the appendix). These prompts are inputted to the \textit{LLM-as-a-Judge}, which then generates scores and justifications for each category's evaluation, adopting the prompt structure of \cite{yuan2024selfrewarding}. To assess \textit{Relevancy}, \textit{Application Specificity}, and \textit{Grammar}, the LLM primarily considers the review and the model's prediction, without referencing the expert's answer. However, for \textit{Accuracy}, the LLM  does reference the expert-provided ground truth answer (i.e.,$\{\textit{ExpertResponse}[R_{i}]\}_{i=1}^{N}$). To further improve the accuracy assessments by the LLM, we integrate the knowledge base of our application and implement the RAG pipeline outlined in \ref{sec:RAG}, aiming to make the judgments more credible and precise. It is worth noting that to deploy our \textit{LLM-as-a-Judge} in real-time, where human expert responses are unavailable, one should omit the human expert response from the evaluation prompts.
\subsection{Iterative Refinement of Customer Review Response}
Once the LLM has demonstrated its capability to assess customer review responses with accuracy comparable to human evaluators, we utilize our optimized LLM as a judge utility to enhance the quality of response  generation flow. This time around, we iterate on only the \textit{M}  which represents the reviews with the lowest scores from human evaluation, indicating areas where improvement is needed. To prevent overfitting, we also include a small proportion of randomly selected reviews in this subset. We refer to this curated set of reviews, where the response generation process has not performed optimally, as 
\begin{align}
\{\text{IR}_{j}\}_{j=1}^{M} \subset \{R_{i}\}_{i=1}^{N}
\end{align}
when $IR$ stands for improvement required. We iterate until the score meets the quality criteria  or we reach a fixed point. 
\begin{align}
\textbf{Judge}(IR_{j}) \geq 0.95  
\end{align}
At each iteration, we modify the prompt guiding the LLM for response generation, instructing it to enhance its performance by utilizing insights from the human expert's answer. This adaptive strategy is designed with the aim that such improvements will be applicable more broadly to the generation of responses for future reviews.
\subsection{LLM as a Response Generator}
\begin{figure}
    \centering
    \includegraphics[width=\linewidth]{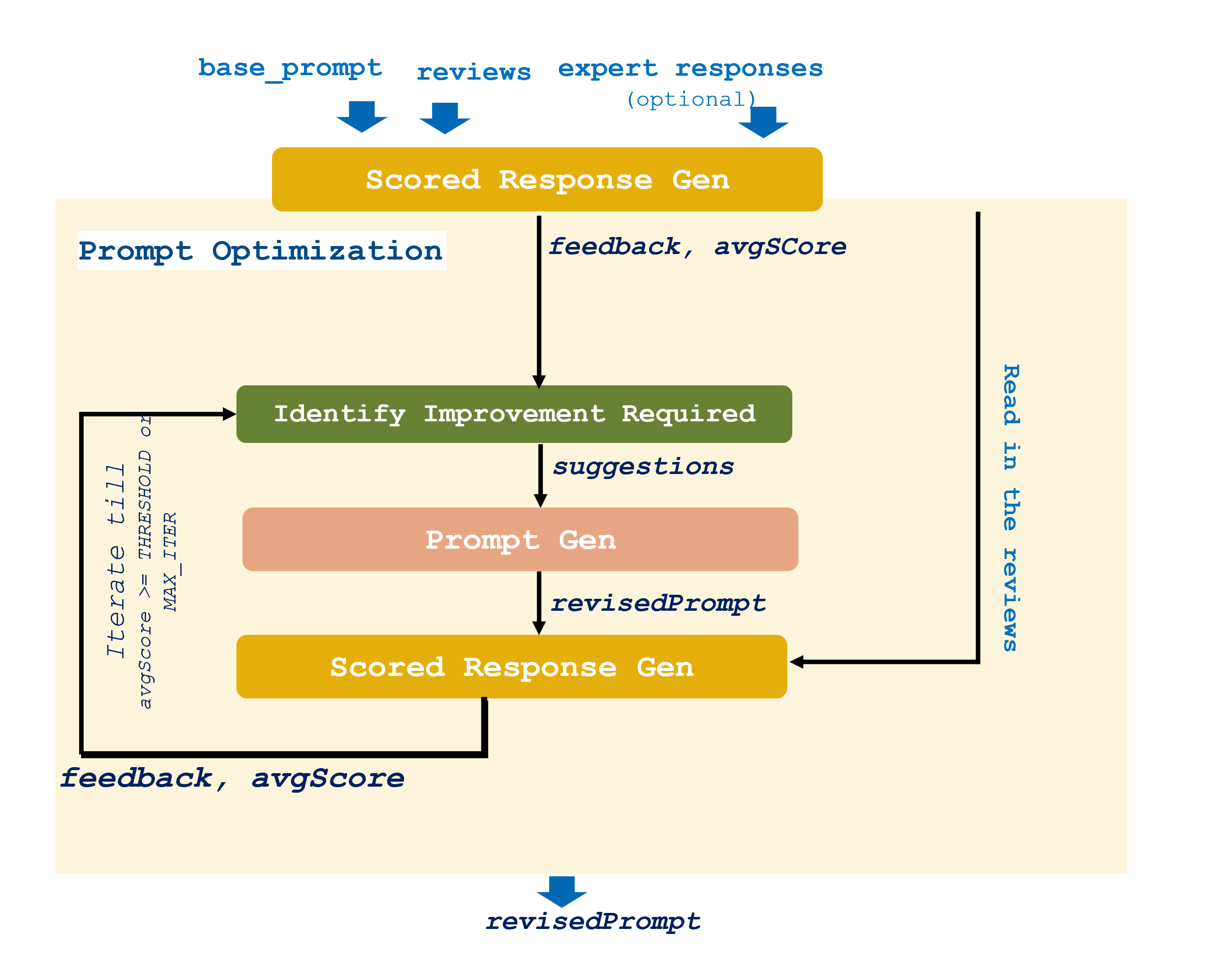}
    \caption{Prompt Optimization Flow driven via feedback of LLM as a judge utility} 
    \label{fig:imgGetResponse}
\end{figure}

\begin{figure}[!ht]
    \centering
    \includegraphics[width=\linewidth]{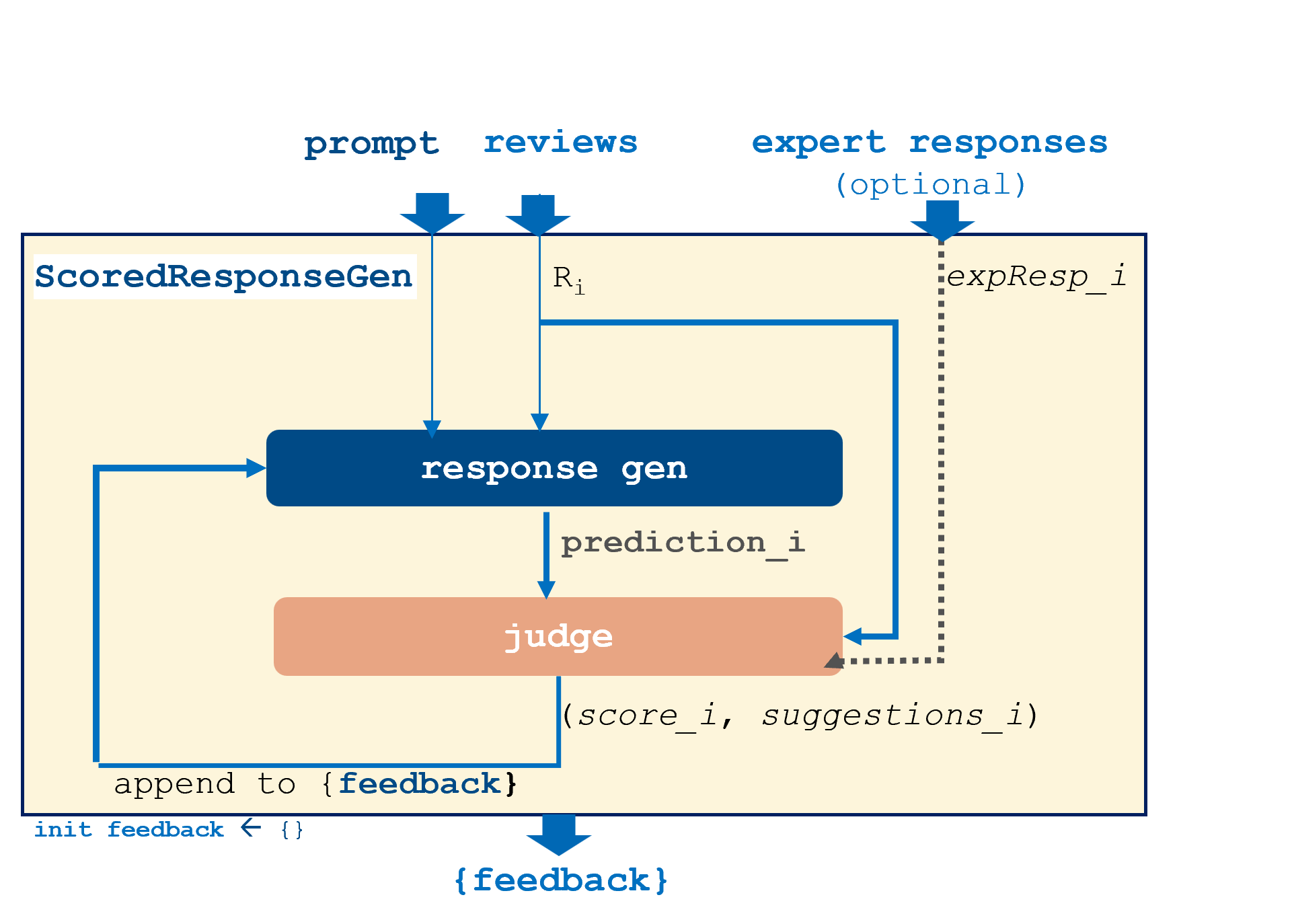}
    \caption{\textbf{ScoredResponseGen} : Given reviews of interest, a prompt and optionally respective expert responses, LLM predicts responses via (\textit{ResponseGen}) utility. Scoring of the quality of the response and  improvement suggestions are handled via (\textit{Judge}) utility. The feedback ouput is a list of score and suggestions pairs for each review. } 
    \label{fig:imgGetFeedback}
\end{figure}

Our iterative self improving flow illustrated in Algorithm \ref{algo:PromptOpt} and Figure  \ref{fig:imgGetResponse} initially generates response predictions for all \text{N} reviews $\{R_{i}\}_{i=1}^{N}$ via instructing LLM (in our case GPT4)  via  the base prompt. The predictions are then evaluated, with scores and feedback, including suggestions for improvements being collected for all reviews as depicted in Algorithm  \ref{algo:ScoredResponse} and Figure \ref{fig:imgGetFeedback}. This results in a collection of (\textit{score}, \textit{suggestions}) tuples for all reviews. Reviews with lowest scores (those for which an improvement is required) are flagged. For these reviews, feedback is specifically sought to refine the responses; denoted as $\{IR_{j}\}_{j=1}^{M}$. Following the refinement of the prompt, we calculate the average score for new response predictions across all reviews using this updated prompt. The process is repeated until  no further improvements may be achieved or we have reached the quality threshold. The end product of this self improving iterative flow is a customized optimized prompt (i.e., \textit{revisedPrompt})  that yields the highest score for customer review response predictions through GPT).  We adopt a dual strategy approach \: \( 1\) Aim to improve reviews with the most need for improvement by selecting \textit{n} \% (in our case $n=30$) of the lowest scoring responses (based on \textit {judge} scores; \( 2\) Incorporate stochasticity to combat overfitting via targeting to improve \textit{m}\%  (in our case $m=10$) additional random response predictions. 
\begin{algorithm}[!ht] 
\caption{Customer Service Chatbot Assistant - Iteratively self  improving customer review response generation based on feedback }    
\label {algo:PromptOpt}
\begin{algorithmic}[1]   
\STATE prompt $\gets$ Basic Prompt Template
\STATE reviews $\gets$ $\{R_{i}\}_{i=1}^{N}$
\STATE feedback $\gets$ \textbf{ScoredResponseGen}(prompt, reviews)
\STATE \textcolor{gray}{\#} \textit{\textcolor{gray}{a list of score \& suggestion pairs for each review}} 
\STATE avgScore $\gets$ \textbf{AverageScore}(feedback)  
\REPEAT
\STATE \text{suggestions} $\gets$ \textbf{IdentifyIR}(feedback) 
\STATE  \textcolor{gray}{\#} \textit {\textcolor{gray}{a list of suggestions  for improvement}}
\STATE  \textcolor{gray}{\# \textit {for \%lowest scoring response predictions}}
\STATE  \textcolor{gray}{\# \textit {and \% of random predictions}} 
\STATE \textcolor{blue}{\textbf{prompt}}  $\gets$  \textcolor{green}{\textbf{PromptGen}}(\textcolor{blue}{\textbf{suggestions}})  
\STATE feedback $\gets$  \\ \textbf{ScoredResponseGen}(prompt, reviews)
\STATE \textcolor{gray}{\#} \textit{\textcolor{gray}{feedback for  response gen. via the revised prompt}}
\STATE avgScore $\gets$  \textbf{AverageScore}(feedback)
\UNTIL $(\text{avgScore} \geq \text{THRESHOLD})$ \text{ or } \text{MAX\_ITER} 
\RETURN \textcolor{blue}{\textbf{prompt}}    
\end{algorithmic}    
\end{algorithm}  
As can be seen in Algorithm \ref{algo:ScoredResponse} and respective Figure \ref{fig:imgGetFeedback}, we leverage LLM both as a generator of response predictions (i.e., \textcolor{green}{\textbf{ResponseGen}}), and also judge the quality of the predictions according to four categories \textit{Relevancy}, \textit{Application Specificity}, \textit{Accuracy} and \textit{Grammatical Correctness} (i.e., \textcolor{green}{\textbf{Judge}}).
\begin{algorithm} [!ht]    
\caption{ Get scored response predictons for the reviews at hand including improvement suggestions for each}   
\label {algo:ScoredResponse} 
\begin{algorithmic}[1]   
\STATE \textbf{function} \textbf{ScoredResponseGen}(prompt, reviews)
\STATE \textcolor{gray}{\#}  \textit{{\textcolor{gray}Generate scored response predictions} }  
\STATE  \textcolor{blue}{\textbf{feedback}}   $\gets$ Empty List   
\STATE i $\gets$ 1  
\FOR{each $R_{i}$  in  reviews}  
\STATE $\textit{prediction}_{i}$  $\gets$ \textcolor{green}{\textbf{ResponseGen}}($R_{i}$)  
\STATE
\begin{align*}  
&(\textit{score}_{i}, \textit{suggestions}_{i}) \\  
&\gets \textcolor{green}{\textbf{Judge}} (R_{i},  \textit{prediction}_{i}, \textit{ExpertResponse}[R_{i}])    
\end{align*}  
\STATE append ($\textit{score}_{i}$, $\textit{suggestions}_{i}$) to \textcolor{blue}{\textbf{feedback}}    
\STATE $i++$  
\ENDFOR  
\RETURN \textcolor{blue}{\textbf{feedback}} 
\end{algorithmic}    
\end{algorithm}  
\subsection{LLM as a Prompt Generator} 
The process of refining prompts through LLM is preceded by a rigorous selection of inputs. Initially, all responses generated by the initial prompt are inspected, and only those with the lowest average scores are chosen for further analysis, as detailed in the previous section. To ensure focused improvement, each response category is further filtered to include only those areas where performance falls below a specific threshold, indicating considerable room for improvement. After the selection has been refined, the LLM embarks on the optimization phase, where it reassesses the original agent's prompt within the context of the selected analyses. The aim here is to enhance clarity, eliminate redundancy, and focus on rectifying the identified weaknesses. This custom-made optimization ensures that the most crucial areas of communication are addressed, thereby augmenting the effectiveness of future responses. By focusing on the response's most critical points, the refined prompt is engineered to bolster the system's overall performance. This vital stage in the continuous loop of prompt optimization also acts as a safeguard against overfitting. It transforms a compilation of specific case responses into concise, actionable prompt instructions that can be generalized across various interactions.
\begin{figure}[!ht]
\begin{center}
\includegraphics[scale=0.5, width=\linewidth]{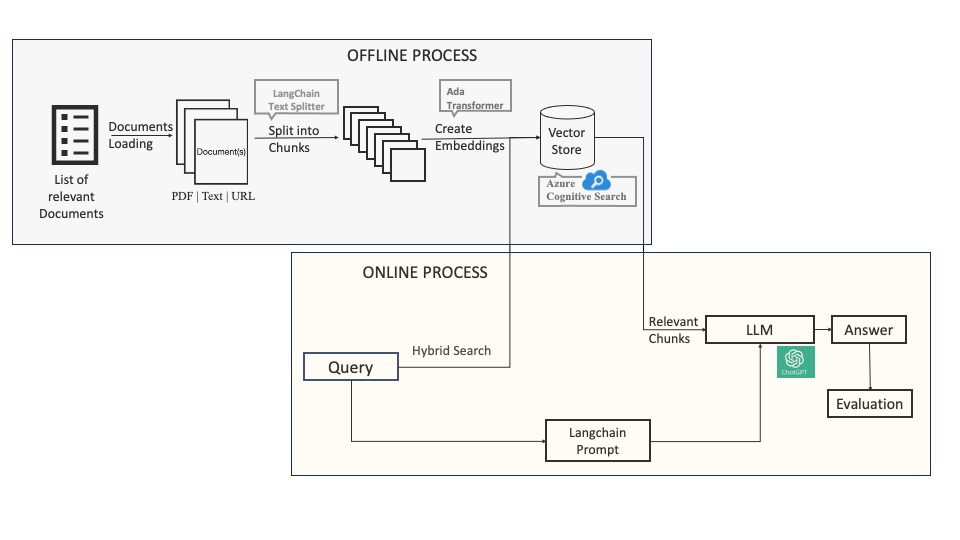}
    \caption{Our Retrieval Augmented Generation Pipeline}
    \label{fig:img1}
\end{center}
\end{figure}
\subsection{Offline and Online Information Retrieval}
\label{sec:RAG}
In the RAG system, as can be seen in Figure \ref{fig:img1}, two distinct yet interconnected workflows, offline and online, are utilized to provide a seamless information retrieval and response generation process. \\
 \textbf{Offline Flow:} The offline flow is dedicated to preparing and structuring the data for the RAG system. This involves the following series of steps:
\begin{enumerate}
\item \textit{Document Loading:} A variety of app-related documents are imported into the system using LangChain Document Loaders that support multiple formats.
\item \textit{Document Segmentation:} Through the LangChain Character Text Splitter, documents are segmented into 500-token pieces to facilitate easier model interpretation.
\item \textit{Generating Embeddings}: The OpenAI text-embedding-ada-002 model is employed to transform document segments into embeddings for better comparison capability.
\item \textit{Storing Documents and Embeddings:} Finally, documents and embeddings are securely stored in a vector store, with Azure AI Search providing straightforward retrieval.
 \end{enumerate}
\textbf{Online Flow:} The online flow is activated when the system interacts with a user's query. It employs a dynamic approach:
\begin {enumerate}
\item \textit{Hybrid Retrieval:} Using Azure Cognitive Search, the system retrieves the top four segments most relevant to the user's query.
\item \textit{Response Generation:} GPT-4 integrates the query with the retrieved information to craft a comprehensive and contextually accurate response.
By combining these offline and online methods, the RAG system ensures the provision of relevant, accurate, and app-specific responses, including useful references and links, in real-time, leveraging both the vast indexed knowledge and the generative capabilities of the advanced AI model.
 \end{enumerate}
\section{Experiments and Results}
\label{sec:results}
In this section, we provide detailed information about our experiments and their corresponding results. The findings from our study suggest that employing a GPT4 LLM can effectively:
\begin{itemize}
\item  Generate automatic responses to customer reviews.
\item Achieve good (close to human)  evaluations of the quality of customer review responses.
\item Automate the enhancement of the LLM's ability to generate responses to customer reviews, ultimately competing with outcomes obtained from human-optimized prompts.
\end{itemize}
\subsection{Customer Review Data}
We collected forty nine real-life customer reviews pertaining to \textcolor{red}{\textless OUR APP NAME \textgreater}\footnote {Application name has been left out} in addition to expert responses from various online platforms, and then split them into train ($28$ reviews) and test ($21$ reviews) datasets. Additionally, we created an extensive knowledge base that includes the application's documentation, such as user manuals and instructional guides to be used in our RAG flow.
\subsection{Human Evaluation}
Analogous to the methodology employed by \cite{bhaskar2023prompted}, the authors of the present study  were tasked with evaluating responses to customer reviews that were produced by a manually refined prompt. Our focus was targeted on various key aspects, namely \textit{Relevancy}, \textit{Application Specificity}, \textit{Accuracy}, and \textit{Grammatical Correctness}. The authors received detailed instructions on how to rate each category separately. The scores given by the human judges are compiled in Table \ref{table:human_scoring}, which includes metrics such as Krippendorff's alpha and Fleiss kappa. Ultimately, the average score for each category, as determined by the labelers, was calculated and normalized to a $0-1$ scale using the min-max normalization.

\begin{table*}[!ht]
\begin{center}
\resizebox{\textwidth}{!}{
\begin{tabular}{ |c|c|c|c|} 
\hline
 Category & Krippendorff's Alpha & Fleiss Kappa & Mean $\pm$ Std\\
\hline
App Specificity & $0.13$ / $0.10$ & $0.05$ / $0.02$ & $4.61 \pm 0.76$ / $4.71 \pm 0.59$\\ 
\hline
Accuracy & $0.26$ / $0.44$ & $0.15$ / $0.11$ & $3.67 \pm 1.11$ / $3.69 \pm 1.24$\\ 
\hline
Relevancy & $0.17$ / $0.2$ & $0.21$ / $0.05$ & $4.90 \pm 0.35$ / $4.83 \pm 0.48$\\ 
\hline
Grammatical Correctness & $-0.01$ / X & $-0.02$ / X & $4.98 \pm 0.13$ / $5.00 \pm 0.00$\\ 
\hline
\end{tabular}
}
\caption{Train/Test Sets - Human Scores}
\label{table:human_scoring}
\end{center}
\end{table*}
\subsection{LLM as a Judge}
\begin{table*}[!ht]
\begin{center}
\resizebox{\textwidth}{!}{
\begin{tabular}{ |c|c|c|c|c|c|c| } 
\hline
 Category & Kendall's $\tau$ & Pearson Correlation& Spearman Correlation& $l_1$ &  $l_2$  & $l_{\infty}$\\
\hline
Relevancy & $0.23$ / $-0.24$ & $\mathbf{0.46}$ / $-0.16$ & $0.24$ /$-0.25$ & $0.67$ / $1.05$ & $0.28$ / $0.42$ & $0.19$ / $0.31$\\ 
\hline
Accuracy & $0.51$ / $\mathbf{0.35}$ & $0.65$ / $\mathbf{0.49}$ & $\mathbf{0.67}$ / $\mathbf{0.49}$ & $3.65$ / $3.69$& $0.94$ / $1.06$& $0.44$ / $0.63$\\ 
\hline
App Specificity & $0.47$ / $-0.23$ & $\mathbf{0.82}$ / $-0.28$ & $0.54$ / $-0.24$ & $2.29$ / $1.69$ & $0.68$ / $0.52$ & $0.50$ / $0.25$\\ 
\hline
Grammatical Correctness & $-0.05$ / X & $-0.05$ / X & $-0.05$ / X & $\mathbf{0.17}$ / $\mathbf{0.10}$ & $\mathbf{0.10}$ / $\mathbf{0.06}$ & $\mathbf{0.08}$ / $\mathbf{0.05}$\\ 
\hline
Overall & $0.39$ / $0.31$ & $0.30$ / $0.46$ & $\mathbf{0.50}$ / $0.43$ & $2.77$ / $1.40$& $0.78$ / $0.38$ & $0.42$ / $0.19$\\ 
\hline
\end{tabular}
}
\caption{LLM-as-a-Judge Compared to Human Scores - Train/Test Sets}
\label{table:llm_judge}
\end{center}
\end{table*}
\label{sec:LLM-as-a-Judge}
Like the human assessment process, the scores from \textit{LLM-as-a-Judge} are also normalized. It is important to note, however, that while the "overall score" from human evaluations is an average of the four categories (after normalization), our observations indicated that placing additional emphasis on the accuracy aspect made the LLM's overall scores more aligned with effective outcomes. Thus, the "overall" score of the \textit{LLM-as-a-Judge} is computed by a weighted average of the categories, with accuracy being given twice the weight of the other categories. A comparison of our LLM and human scores is presented in Table \ref{table:llm_judge}. Within the training data set, our \textit{LLM-as-a-Judge} shows moderate to strong positive correlation with human scores in the same category in three categories (\textit{Relevancy}, \textit{Accuracy}, and \textit{Application Specificity}) and in the overall score. The fourth category, which presents nearly zero correlation, still exhibits a negligible variance between the LLM and human scores. Moreover, only a few human scores in the Grammar category are less than $5$, high grammatical quality generation by GPT-4. For the test dataset, the \textit{Accuracy} and the overall scores  moderately correlate to those from humans, paired with a nearly exact match in \textit{Grammatical Correctness}. We note that that for the test set, all human scores were at the $5$, thus calculations  of Krippendorff's Alpha and Fleiss Kappa are irrelevant. However, unlike the training set, the \textit{Relevancy} and \textit{Application Specificity} scores of the LLM showed a weak (and negative) correlation with human assessments. To demonstrate the strength of our \textit{LLM-as-a-Judge} we compared the overall score obtained using our evaluation prompts and the prompt of \cite{yuan2024selfrewarding} against the human grades (Table \ref{table:prompt_compare}). Our experiments imply that a tailored evaluation prompt for each category, specifically related to customer support, is more advantageous than a single broad evaluation prompt. To make the comparison as fair as possible, we made few changes to the original prompt of \cite{yuan2024selfrewarding}. First, we made the prompt more suitable to customer review domain, for example, we replaced the word 'question' with the word 'review'. We also add the product context to the prompt, similarly to our prompt, enhancing the judge capabilities. Lastly, we tested how adding a reference to the ground truth expert response, affect the scores. The assessment was conducted by calculating the correlation and divergence between these LLM-assigned scores and the scores obtained from human assessments of responses generated by the manually optimized prompt.
\begin{table*}[!ht]
\begin{center}
\resizebox{\textwidth}{!}{
\begin{tabular}{ |c|c|c|c|c|c|c| } 
\hline
 Category & Kendall's $\tau$ & Pearson Correlation& Spearman Correlation& $l_1$ &  $l_2$  & $l_{\infty}$\\
\hline
Overall - \cite{yuan2024selfrewarding} & $0.10$ / X & $0.13$ / X & $0.12$ / X & $4.55$ / $3.15$& $0.95$ / $0.78$ & $\mathbf{0.33}$ / $0.25$\\ 
\hline
Overall - \cite{yuan2024selfrewarding} + Expert Response & $0.07$ / $-0.10$ & $0.08$ / $-0.12$ & $0.09$ / $-0.14$& $7.35$ / $6.55$ & $1.75$ / $1.83$ & $0.89$ / $0.94$\\ 
\hline
\textbf{Overall - Ours} & $\mathbf{0.39}$ / $\mathbf{0.31}$ & $\mathbf{0.30}$ / $\mathbf{0.46}$ & $\mathbf{0.50}$ / $\mathbf{0.43}$ & $\mathbf{2.77}$ / $\mathbf{1.40}$ & $\mathbf{0.78}$ / $\mathbf{0.38}$ & $0.42$ / $\mathbf{0.19}$\\ 
\hline
\end{tabular}
}
\caption{LLM-as-a-Judge Prompt Comparison : Train/Test Sets}
\label{table:prompt_compare}
\end{center}
\end{table*}
\begin{table*}[!ht]
\begin{center}
\resizebox{\textwidth}{!}{
\begin{tabular}{ |c|c|c|c| } 
\hline
Category & LLM Scoring (Base) & LLM Scoring (Human Optimized) & LLM Scoring (LLM Optimized) \\
\hline
App Specificity& $0.76$ & $0.93$ & $\mathbf{0.99}$ \\ 
\hline
Accuracy & $0.72$ & $0.78$ & $\mathbf{0.84}$ \\ 
\hline
Relevancy & $0.94$ & $\mathbf{0.99}$ & $0.97$ \\ 
\hline
Grammatical Correctness & $0.98$ & $1.00$ & $1.00$ \\ 
\hline
Overall & $0.81$ & $0.87$ & $\mathbf{0.91}$ \\ 
\hline
\end{tabular}
}
\caption{LLM Scores of Generated Responses - Train Set}
\label{table:response_gen_res_train_score}
\end{center}
\end{table*}
\begin{table*}[!ht]
\begin{center}
\resizebox{\textwidth}{!}{
\begin{tabular}{ |c|c|c|c| } 
\hline
Category & LLM Scoring (Base) & LLM Scoring (Human Optimized) & LLM Scoring (LLM Optimized) \\
\hline
App Specificity& $0.92$ & $0.99$ & $0.99$ \\ 
\hline
Accuracy & $0.78$ & $0.79$ & $\mathbf{0.81}$ \\ 
\hline
Relevancy & $0.99$ & $0.99$ & $0.99$ \\ 
\hline
Grammatical Correctness & $0.99$ & $1.00$ & $1.00$ \\ 
\hline
Overall & $0.87$ & $0.89$ & $\mathbf{0.90}$ \\ 
\hline
\end{tabular}
}
\caption{LLM Scores of Generated Responses - Test Set}
\label{table:response_gen_res_test_score}
\end{center}
\end{table*}

\subsection{LLM as a Response Generator}
\begin{figure*}[!ht]
    \centering
        \includegraphics[width=\linewidth]{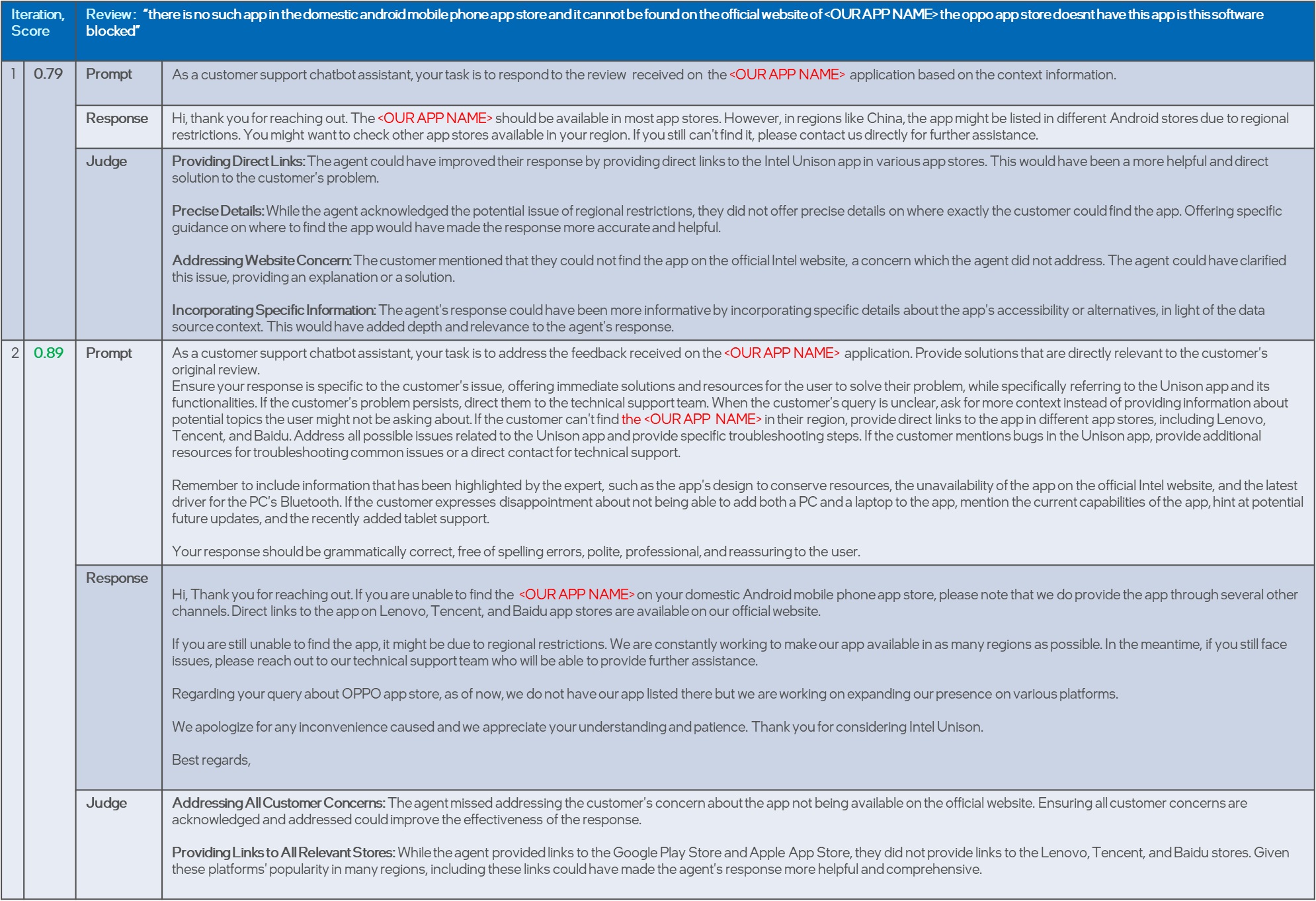}
    \caption{Iterative Self-Improving  Response Generation Step}
    \label{table:ResponseGen}
\end{figure*}

\begin{table*}[!ht]
\begin{center}
\resizebox{\textwidth}{!}{
\begin{tabular}{ |c|c|c| } 
\hline
Category & (Normalized) Averaged Human Scoring (Base) & (Normalized) Averaged Human Scoring (LLM Optimized) \\
\hline
App Specificity& $0.77$ & $0.87$ ($+\mathbf{12.99\%}$) \\ 
\hline
Accuracy & $0.60$ & $0.68$ ($+\mathbf{13.33\%}$)\\ 
\hline
Relevancy & $0.76$ & $0.84$ ($+\mathbf{10.53\%}$)\\ 
\hline
Grammatical Correctness & $1$ & $1$ \\ 
\hline
Overall & $0.78$ & $0.85$ ($+\mathbf{8.97\%}$)\\ 
\hline
\end{tabular}
}
\caption{Human Scores of Generated Responses - Test Set}
\label{table:test_set_human_scores}
\end{center}
\end{table*}
Our study utilized various examples to showcase the strength of our refined response generation mechanism. Initially,  Tables \ref{table:response_gen_res_train_score} and \ref{table:response_gen_res_test_score} illustrate that the outputs crafted using our LLM-tailored prompts outperform responses generated with human-tailored and foundational prompts in almost every aspect. This superior performance is consistently observed across both train and test datasets, as evaluated by our LLM-as-a-Judge. Further, in Figure \ref{table:ResponseGen} we provide an insight on the improvements obtained in each iteration of our self-improving response generation flow via providing details of prompt, response, score and suggestions of the LLM for an iteration step. To impartially assess the improvement in the results achieved using the base prompt versus our optimized prompt, we enlisted four team members, unaffiliated with this project, to manually score the test set generation obtained using the base and optimized prompts. The scorers were kept blind to the origin of the results, i.e. which were derived from which prompts. Although the LLM exhibited a slight improvement with the refined prompt, Table \ref{table:test_set_human_scores} reveals a significantly larger improvement regarding to human scoring of more than $8.5\%$ overall. Finally, we generated responses for $50$ new reviews and solicited a domain expert to evaluate the results, aiming to gain a general understanding of the result quality for new, unseen examples.  The scores obtained were noteworthy, averaging $4.68$ for relevancy, $4.8$ for accuracy, $4.7$ for application specificity and $4.32$ for grammatical correctness.
\section{Discussion} 
\label{sec:discussion}
Building on prior research in the domain of customer review response creation, our study integrates state-of-the-art machine learning technologies, particularly LLMs. We present a novel contribution with our \textit{LLM-as-a-Judge}, an automated evaluation method to assess customer review responses (vs. ground truth). Our findings support the use of tailored evaluation prompts for each review category over the application of a single, more generic prompt. The data indicates that responses crafted with our refined prompt align closer to human responses by $3-5$ fold in terms of correlation. For practical application, the refined prompt can be implemented at a production level. Considering the frequent updates to customer support materials and databases, we recommend regular refreshes to provide the latest data for the RAG, thereby reducing inaccuracies in the model's outputs. In parallel, to adapt to the continual influx of customer reviews, we advocate for regular retraining of the model to derive new and improved prompts. Another insight of our research is the potential utilization of comparison of LLM and human evaluation scores to let us understand when new knowledge (data points) need to be added to our input knowledge (i.e., RAG) pipeline. Should the LLM as a judge score fall below the human evaluation score, it indicates that the LLM can learn how to improve by referencing the human expert's response. Conversly, if the LLM as a judge significantly exceed the human evaluation score; i.e., by at least 0.1, we may assume that  our  LLM based  response generation lacks the needed knowledge to improve and request LLM to create generalized new data points (i.e., Q\&A  data points) leveraging the review and human expert response. These newly created data constructs can then be reincorporated into our generation process to enhance the quality of responses for future reviews.
\section{Conclusions}
\label{sec:conclusions}
In summary, our comprehensive preparation of customer review data for both training and testing, combined with the utilization of human evaluators, has enabled us to thoroughly assess the ability of the LLM (GPT4 in particular) to act as an effective response generator to customer reviews of  \textcolor{red}{\textless OUR APP NAME \textgreater} at an app store. Our experimental results provide strong evidence of LLMs dual functionality. Not only can they effectively generate predictive responses to customer reviews, but they also show a commendable capacity to evaluate the quality of those response predictions. This dual functionality enhances the system's adaptability and versatility, making it a valuable tool in the realm of customer service and communication. The outcomes from our assessments provide a promising foundation for further exploration and improvement of LLMs capabilities in practical real-world settings.

\section{Acknowledgments}

We wish to express our sincere gratitude to Nady Gorovetsky and Tomer Achdut for their invaluable contributions in this work. Their expert responses to customer reviews and insightful feedback were instrumental. This research greatly benefited from their knowledge and dedication. Their involvement has been a significant factor in the progress of this project. 

\nocite{*}
\clearpage
\section{Bibliographical References}\label{sec:reference}

\bibliographystyle{scrable-natbib}
\bibliography{Bibliography-2}

\bibliographystylelanguageresource{scrable-natbib}
\bibliographylanguageresource{languageresource}

\appendix

\section{Appendix}
\subsection{Prompts}\label{appendix:prompts}
\begin{tcolorbox}[title=\textit{Base Prompt}, label={BasePrompt}]
Instruction: \\
As a customer support chatbot assistant, your task is to respond to the review received on the \textcolor{red}{\textless OUR APP NAME \textgreater} application based on the context information. \\
Context: {context} \\
Question: {question} \\
Answer: \\
\end{tcolorbox}

\begin{tcolorbox}[title=\textit{Human Optimized Prompt}, label={HumanOptimizedPrompt}]
As a customer support chatbot assistant, your task is to respond to the review received on the \textcolor{red}{\textless OUR APP NAME \textgreater} application. 
Your goal is to craft a response that will satisfy and delight the customer who wrote the review. 
Please follow the steps below:
\begin{enumerate}
	\item Analyze the customer's question and the context provided.
	\item Formulate a response that addresses their concerns or queries.
	\item Only for the issues that may necessitate professional intervention, please include this message in your response: 'Should the problem continue, we encourage you to contact our technical support team for expert help. [support url]'
    \item If the context contains useful information that can assist the user, incorporate it into your response, add helpfull links from the context, if link is added do not add the same link again as a reference.
    \item In case the context does not provide any relevant information, use your general knowledge to formulate a helpful response.
	\item Start your response by thanking user for their feedback, and ensure that your response is short and highlights the positive features of the \textcolor{red}{\textless OUR APP NAME \textgreater} application.
\end{enumerate}      
Context: \{context\} \\
Question: \{question\} \\
Answer: \\
\end{tcolorbox}

\begin{tcolorbox}[title=\textit{LLM Optimized Prompt}, label={LLMOptimizedPrompt}]
Instruction: \\
As a customer support chatbot assistant, your role is to respond to the feedback received about the \textcolor{red}{\textless OUR APP NAME \textgreater} application. Tailor your responses to the customer's specific issue, offering helpful solutions and resources.\\
Context: \{context\}\\
Customer review: \{question\}\\
Answer:\\
In your response, ensure you address the customer's primary issue and provide immediate, actionable solutions. Refer to the \textcolor{red}{\textless OUR APP NAME \textgreater} app and its features, and guide them to the technical support team if the issue persists. Should the customer's query be unclear, clarify by asking for more information. If the customer can't locate the \textcolor{red}{\textless OUR APP NAME \textgreater} app, provide direct links to different app stores. Address all potential issues related to the \textcolor{red}{\textless OUR APP NAME \textgreater} app by providing clear troubleshooting steps. If the customer mentions bugs in the \textcolor{red}{\textless OUR APP NAME \textgreater} app, direct them to resources for common troubleshooting or provide contact information for technical support.\\
Remember to highlight key information such as the app's design to conserve resources, its unavailability on certain platforms, and recent updates. If the customer expresses disappointment about certain app capabilities, acknowledge their feedback, explain the current app capabilities, and hint at future updates if applicable. Also, don't forget to mention the recently added tablet support.\\
Your response should be grammatically correct, free of spelling errors, and maintain a polite and professional tone. Use the data source context effectively without being overly lengthy or repetitive. Focus on directly addressing the user's review and providing a concise, relevant response.
\end{tcolorbox}

\begin{tcolorbox}[title=\textit{LLM Prompt Optimization}, label={LLMPromptBuild}]
Your task is to enhance the effectiveness of customer service interactions. Begin by reviewing the original agent's prompt and the analysis of the responses it generated. \\
Use the insights from the analysis to refine the agent's prompt, aiming to improve the agent's overall performance for future interactions. \\
Your revised prompt should be clear, concise, and non-repetitive. \\
Your revised prompt should be focused on addressing the identified areas for improvement while retaining the structure of the original prompt. \\
Be sure to enclose all variables in curly brackets as in the original prompt. \\
Begin your revision process here: \\
Original Agent's Prompt: \{question\} \\
Responses Analysis: \{context\} \\
Your Improved Prompt: \\
\end{tcolorbox}

\begin{tcolorbox}[title=\textit{LLM-as-a-Judge} - Accuracy Prompt, label={accPrompt}]

You get a customer review, a corresponding customer service agent response, 
and a best possible response designed by an expert. Your role is to rate 
how accurate the agent's response, based on the context and the expert response. 
Your score should be based on the following criteria - Accuracy -
\begin{enumerate}
    \item Based on the expert's response, does the agent's response answer the user concerns regarding to the \textcolor{red}{\textless OUR APP NAME \textgreater} app accurately?
    \item Does the agent's response lack some information from the expert's response?
    \item Does the agent's response aligned with the expert response? 
    \item Does the agent use the Data Source Context correctly to generate the answer? 
    \item Does the agent use the Data Source Context accurately when addressing the user concerns?
\end{enumerate} 
Assign a score ranging from 1.0 to 5.0, where 1.0 signifies inaccurate response and 5.0 indicates very accurate response. 
Dont refer the quality of the answer, only refer to its accuracy. 
Your output must be a single number between 1.0 to 5.0.\\
Customer review : \{query\} \\
Agent response: \{result\}   \\
Expert Response : \{answer\}\\
After examining the user’s review, the agent's response and the expert's response:
Briefly justify your total score, up to 150 words.
If possible, use the Data Source Context to establish your claims. 
Conclude with the score using the format: 'Total Score: \textless	
total points\textgreater'
\end{tcolorbox}

\begin{tcolorbox}[title=\textit{LLM-as-a-Judge} - Relevancy Prompt, label={relPrompt}]
You get a customer review and a corresponding customer service agent response.
Your role is to rate the relevancy of the agent's response. 
Your score should be based on the following criteria:
\begin{enumerate}
\item Is the response relevant and provides some information related to the user’s review ? 
\item Is the response addressing the user’s review directly? 
\item If not specifically mentioned, you may assume that the user is using the \textcolor{red}{\textless OUR APP FULL NAME \textgreater} app. 
\end{enumerate}
Assign a score ranging from 1.0 to 5.0, where 1.0 signifies a non relevant response and 5.0 indicates a very relevant response. 
Dont refer the quality of the answer, only refer to its relevancy to the user review. 
Your output must be a single number between 1.0 to 5.0.\\    
Customer review : {query}\\
Agent response: {result}\\
Expert Response : {answer}\\   

After examining the user’s review and the agent's response:
Briefly justify your total score, up to 150 words. 
Conclude with the score using the format: 'Total Score: \textless	
total points\textgreater'
\end{tcolorbox}

\begin{tcolorbox}[title=\textit{LLM-as-a-Judge} - Grammatical Correctness Prompt, label={grammarPrompt}]
You get a customer review and a corresponding customer service agent response. 
Your role is to rate the grammar of the agent's response. Your score should be based on the following criteria: 
\begin{enumerate}
    \item Is the response grammatically correct?
    \item Does the response has no spelling errors?
\end{enumerate}    
Assign a score ranging from 1.0 to 5.0, where 1.0 signifies a wrongly spelled, 
low quality response and 5.0 indicates a grammatically correct high quality response. 
Your output must be a single number between 1.0 to 5.0.\\
Customer review : {query}\\
Agent response: {result}\\
Expert Response : {answer}\\

After examining the user’s review and the agent's response:
Briefly justify your total score, up to 150 words. 
Conclude with the score using the format: 'Total Score: \textless	
total points\textgreater'
\end{tcolorbox}

\begin{tcolorbox}[title= \textit{LLM-as-a-Judge} - App Specificity, label={appSpecPrompt}]
You get a customer review and a corresponding customer service agent response. 
Your role is to rate the agent's response regarding whether it specifically addresses to \textcolor{red}{\textless OUR APP FULL NAME \textgreater} app. 
Your score should be based on the following criteria:
\begin{enumerate}
    \item Is the response specifically tailored to \textcolor{red}{\textless OUR APP FULL NAME \textgreater} and its functionalities? 
    \item Do the opening and the end of the response relate to  \textcolor{red}{\textless OUR APP NAME \textgreater}?
    \item If not specifically mentioned, you may assume that the user is using the\textcolor{red}{\textless OUR APP FULL NAME \textgreater}
    \item For your concern, \textcolor{red}{\textless OUR APP NAME \textgreater} and \textcolor{red}{\textless OUR APP FULL NAME \textgreater} are the acronyms.
    \item If not specifically mentioned, you may assume that the user is using the \textcolor{red}{\textless OUR APP FULL NAME \textgreater} app.
    \end{enumerate} 
Assign a score ranging from 1.0 to 5.0, where 1.0 signifies a response that is not specific to \textcolor{red}{\textless OUR APP NAME \textgreater} and 5.0 indicates a response that is very specific for \textcolor{red}{\textless OUR APP NAME \textgreater}.
Dont refer the quality of the answer, only refer to its specifically relates to\textcolor{red}{\textless OUR APP NAME \textgreater}. 
Your output must be a single number between 1.0 to 5.0.\\
Customer review : {query}\\
Agent response: {result}\\
Expert Response : {answer}\\

After examining the user’s review and the agent's response:
Briefly justify your total score, up to 150 words. 
Conclude with the score using the format: 'Total Score: \textless	
total points\textgreater'	
\end{tcolorbox}

\end{document}